\documentclass[11pt,a4paper]{article}
\usepackage[hyperref]{emnlp2020}
\usepackage{times}
\usepackage{latexsym}
\usepackage{graphicx}
\usepackage{soul,color}
\usepackage{subcaption}

\usepackage{microtype}

\aclfinalcopy

\renewcommand\footnotemark{}

\newcommand\ROFT{RoFT}

\title{\ROFT{}: A Tool for Evaluating \\Human Detection of Machine-Generated Text}

\author{Liam Dugan*, \hspace{0.25cm} Daphne Ippolito*, \hspace{0.25cm} Arun Kirubarajan*, \hspace{0.25cm} Chris Callison-Burch\\
\thanks{$^*$Authors listed alphabetically contributed equally.}
University of Pennsylvania\\\
{\tt\small \{ldugan, daphnei, kiruba, ccb\}@seas.upenn.edu}
}
\date{}

\begin{document}
\maketitle
\begin{abstract}
In recent years, large neural networks for natural language generation (NLG) have made leaps and bounds in their ability to generate fluent text.
However, the tasks of evaluating quality differences between NLG systems and understanding how humans perceive the generated text remain both crucial and difficult.
In this system demonstration, we present Real or Fake Text (\ROFT{}), a website that tackles both of these challenges by inviting users to try their hand at detecting machine-generated text in a variety of domains.
We introduce a novel evaluation task based on detecting the boundary at which a text passage that starts off human-written transitions to being machine-generated.
We show preliminary results of using \ROFT{} to evaluate detection of machine-generated news articles.
\end{abstract}

\section{Introduction}
Despite considerable advancements in building natural language generation (NLG) systems that can output extremely fluent English text, there is still not very much understanding of how humans perceive machine-generated text.
Such an understanding is crucial for the evaluation of the improvements in NLG systems and for the analysis of the societal ramifications of machine-generated text as it becomes increasingly easy to produce.

When evaluating NLG systems, it is considered standard practice to ask evaluators to rate generated text on criteria such as fluency, naturalness, or relevance to a prompt on a Likert scale \citep{van2019best}.
Preference studies, where a rater is shown two generated excerpts and asked which one they prefer, are also common.
Some recent work has focused on the detection problem: how capable humans are at distinguishing textual excerpts generated by a system from those written by another human \citep{ippolito2020automatic,zellers2019defending}.

\begin{figure}
    \centering
    \includegraphics[width=0.49\textwidth]{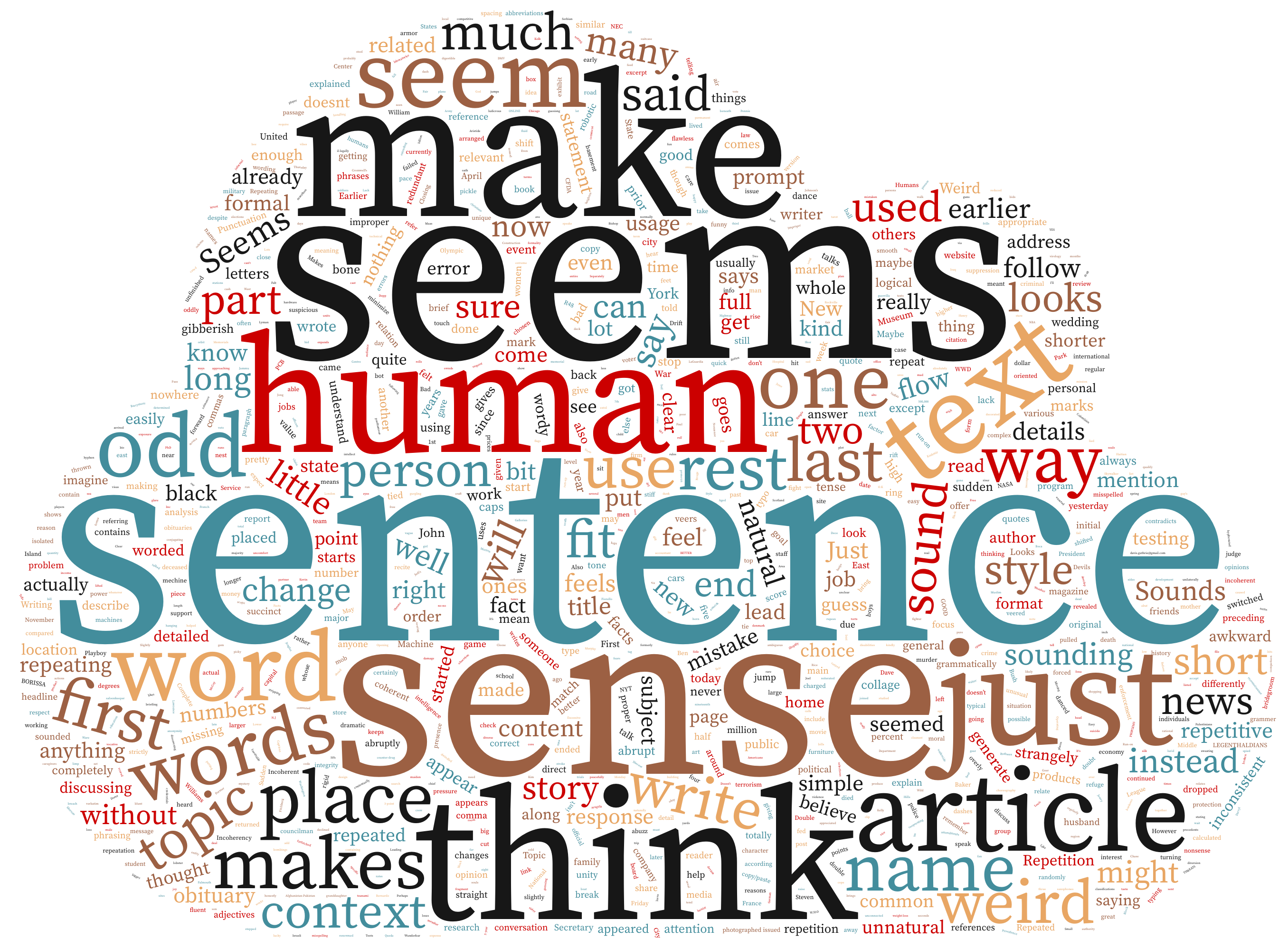}
    \caption{A word cloud of common words that annotators used to describe why they thought sentences were machine-generated.}
    \label{fig:tricklen}
\end{figure}

However, due to the prohibitive cost of running human evaluation studies, most prior  work in this area has been rather limited in scope.
For example, analyses usually show results on only a single category of text (news articles, stories, webtext, etc.).
This could be problematic since different domains have different levels of named entities, world facts, narrative coherence, and other properties that impact the success of NLG systems.
In addition, most papers only evaluate on a very limited selection of decoding strategy hyperparameters. \citet{holtzman2019curious} and \citet{ippolito2020automatic} both show that the decoding strategy chosen at inference time can have a significant impact on the quality of generated text.

In this work, we introduce the Real or Fake Text (\ROFT{}) system, a novel application for simultaneously collecting quality annotations of machine-generated text while allowing the public to assess and improve their skill at detecting machine-generated text.

In \ROFT{}, we propose to use the task of detecting when text is machine-generated as a quality criterion for comparing NLG systems.
Following \citet{ippolito2020automatic}, we make the counterintuitive assumption that the \textit{worse} annotators are at detecting that text is machine-generated, the \textit{better} we can say that the NLG system is at generating text.

In \ROFT{}'s detection task, annotators are shown a passage of text one sentence at a time.
The first several sentences are from a real human-written text source and the next several sentences are a machine-generated continuation. The user's goal is to guess where the boundary is.
When they think that a sentence is machine-generated, they are asked to give an explanation for their choice. Afterwards the true boundary is revealed.

In the remainder of this paper, we discuss why we think this task is interesting from a research perspective and describe the technical details behind our implementation. We show preliminary results that showcase the types of analyses that are possible with the collected data, and finally we discuss plans for future work.

The \ROFT{} website is located at \url{http://www.roft.io/}.
The source code is available under an MIT License at \url{https://github.com/kirubarajan/roft}.

\section{Research Motivations}
\label{section:motivations}
The purpose behind \ROFT{} is to collect annotations on the scale needed to probe the quality of text generated under a variety of NLG conditions and systems.
In this section, we describe three research questions we aim to answer using RoFT data.

\subsection{Length Threshold for Detection}
State-of-the-art generative models tend to produce text that is locally fluent but lacking in long-term structure or coherence. Intuition suggests that fluent NLG systems ought to produce text that is high quality for long durations (measured in number of sentences). As such, we are interested in using the the boundary detection task---whether annotators can detect the boundary between human-written text and a machine-generated continuation---as a comparison method for NLG systems.
We hypothesize that for better quality systems, the generated text will be able to fool humans for more sentences.

\subsection{Text Genre/Style}
Generative language models have now been trained and fine-tuned on a great diversity of genres and styles of text, from Reddit posts \citep{keskar2019ctrl} and short stories \citep{fan2018hierarchical} to Wikipedia \citep{liu2018generating} and news articles \citep{zellers2019defending}.
Each of these datasets has its own distinct challenges for generation; for example, in the story domain it is acceptable for a generator to make up facts while this would be unacceptable in a Wikipedia article.
We are interested in how these differences might impact the ability of humans to detect machine-generated text.

\subsection{Reasons Text is Low Quality}
A study by \citet{van2019best} found that less than 3\% of recent papers on NLG ask for free-text comments when performing human evaluations.
And yet, understanding why humans think text is low quality can be very important for diagnosing problems in NLG systems \citep{reiter2009investigation}.
Therefore, the \ROFT{} platform collects free-form textual explanations from our annotators on their decisions.
Such data, though inevitably noisy, could provide insights into the types of errors that NLG systems introduce, the types of errors humans are sensitive to, and even the types of errors human-written corpora contain (when a rater inadvertently predicts that a human-written sentence is machine-generated).

\subsection{Human Factor}
We do not expect that real-world uses of machine-generated text would involve such a tidy split of prompt sentences followed by a machine-generated continuation.
However, we believe that even an artificial framing such as \ROFT{}'s has both the potential to educate the public on what to look for in machine-generated text and give researchers insights into how humans perceive and react to such text.
We are particularly interested in how annotators may or may not improve over time and in what ways their respective demographics (for example, paid crowd worker vs. university student) impact their detection skill.

\section{System Overview}
This section gives an overview of \ROFT{}'s design, including the task that annotators are asked to complete and methods for encouraging organic traffic.

\subsection{Task Definition}
\begin{figure}
    \centering
    \includegraphics[width=0.5\textwidth]{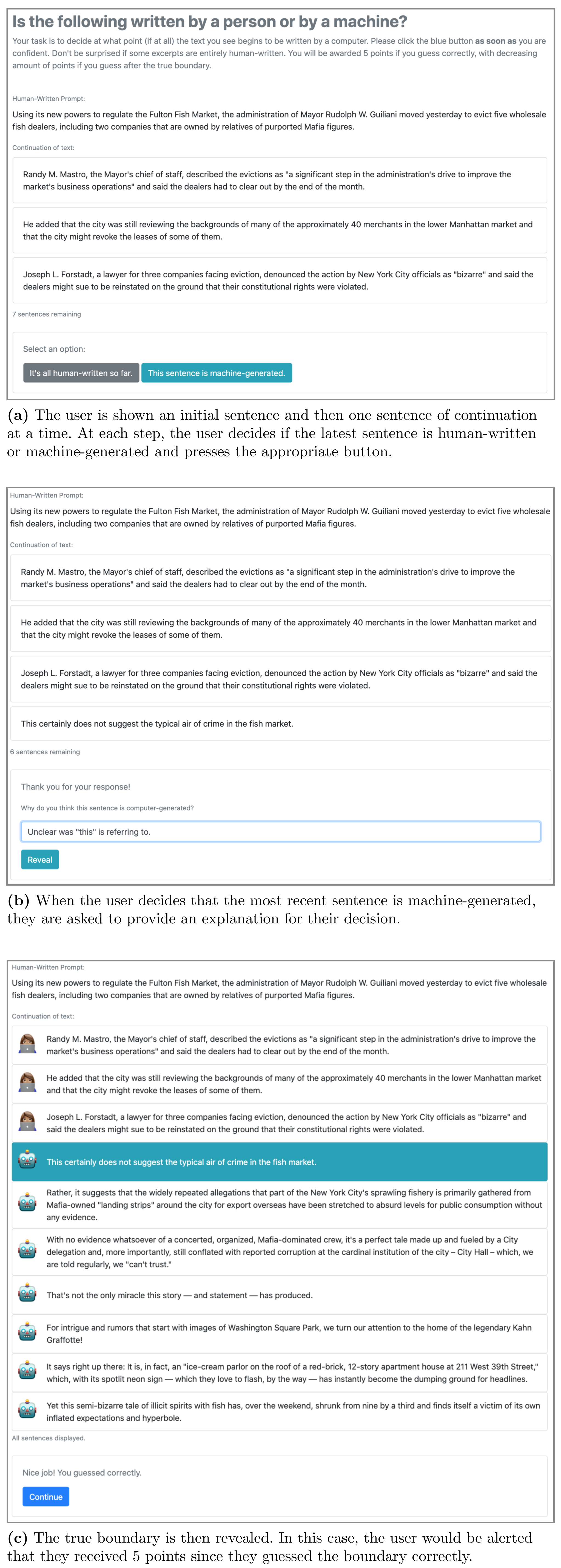}
    \caption{The user interface for annotation.}
    \label{fig:annotation_screenshot}
\end{figure}

The RoFT annotation task is posed as a game.
Users first choose which category they would like to play in (where different categories correspond to different text domains or NLG systems).
The ``game" then consists of a series of rounds.
Each round starts with the user being presented a single sentence that is guaranteed to be human-written.
For example, this might be the first sentence of a New York Times article.
Afterwards, users may select to display more sentences, one at a time.
At each step, they must decide if they believe that the most recent sentence is still written by a human.
When the user decides they are confident that a machine has written the most recent sentence (i.e. they have found the ``boundary sentence"), the round ends.
The user is then asked to provide a natural language explanation of what prompted their decision.
In essence, the annotators' goal is to identify the exact sentence where a machine ``takes over" and the text is no longer human-written.
Figure \ref{fig:annotation_screenshot} gives screenshots of the flow of a single round.

\begin{figure}
    \centering
    \fbox{\includegraphics[width=0.45\textwidth]{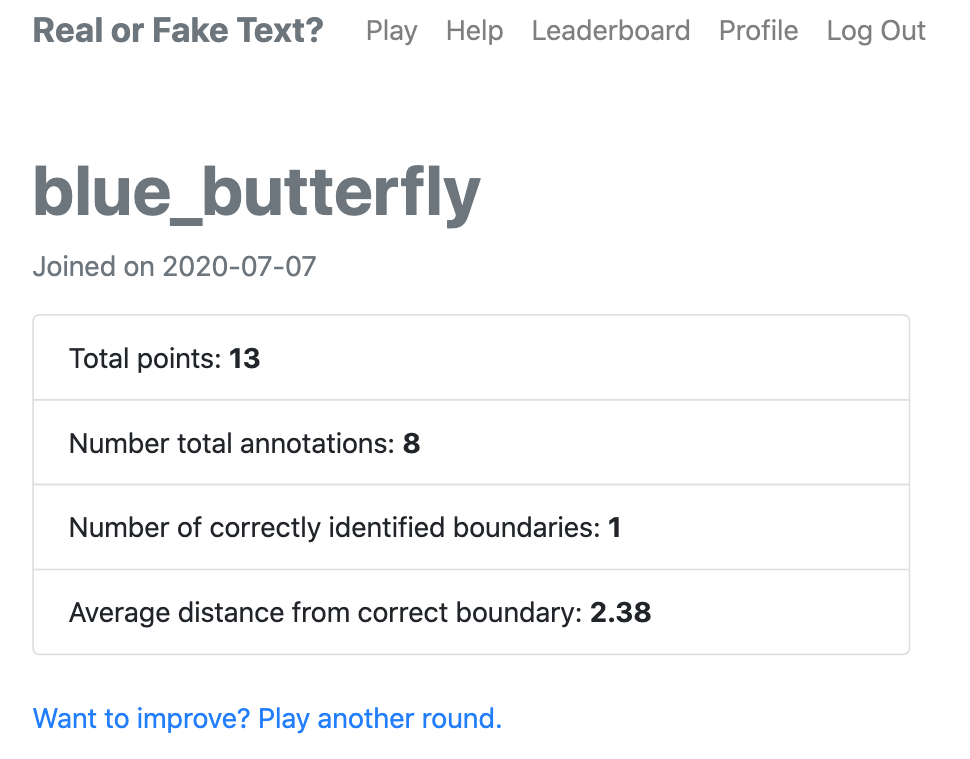}}
    \caption{A user's profile page.}
    \label{fig:profile_Screenshot}
\end{figure}

\subsection{Implementation}
The \ROFT{} annotation website is designed to collect data needed to answer a variety of research questions, including those posed in Section \ref{section:motivations}.
In particular, our system stores detailed metadata for each annotation.
These include the order in which a user completed annotations, the type of user account associated with each annotation (e.g. paid worker or organic traffic), the NLG system used to produce each generation, and the amount of time each annotation took.
The system was developed in Python using the Django Framework and a SQL database.
The use of a relational database enables sophisticated queries to be made on the collected annotations for analysis.
We plan to make dumps of the database available to other researchers to further promote research into the evaluation of generated text.

\subsection{Gamification}
Since the cost of collecting human annotations via a crowd platform such as Amazon Mechanical Turk can be prohibitively expensive for large studies, we aimed to build the \ROFT{} website in a manner that would encourage sustained participation without the need for a financial incentive.

Each user has a Profile page (shown in Figure \ref{fig:profile_Screenshot}) where they can see statistics on the total number of annotations they have done, how many points they have earned, and how many questions they have answered perfectly.
There is also a leaderboard where users can check how their point count compares to other raters.
The leaderboard encourages users to do more annotations, since this is the only way to move up on the rankings.

We received unsolicited compliments from our initial annotators such as ``Interesting, fun task" and ``Really convincing passages." We intend to add further gamification elements, including leaderboards broken down by text domain, comprehensive statistics on user progress and skill, and the ability to see and up-vote the free-text comments of other users.

\subsection{Generations}
We ultimately plan to use \ROFT{} to study differences in detection performance across a variety of NLG systems and text domains.
The initial version of \ROFT{} includes two complementary categories of text: news and fictional stories.
Users have the option to choose which category they would like to annotate.

For the news category, prompts are drawn from the New York Times Annotated Corpus \citep{sandhaus2008new} and are truncated to between 1 and 10 sentences long. GROVER \citep{zellers2019defending} is then conditioned on these starting sentences and asked to complete the article. Finally, the outputs from GROVER are truncated so that the sum total number of sentences for each example is 10.

The data on fictional stories was prepared similarly except that the Reddit Writing Prompts dataset \citep{fan2018hierarchical} was used for the prompts, and the GPT-2 XL model \citep{radford2019language} was used for generation.

Each category contains over 1,500 examples, where for each example the number of human-written context sentences as well as the values of the decoding strategy hyperparameters were chosen randomly. For our initial seeding of data, Nucleus sampling \citep{holtzman2019curious} was used for all decoding, where the $p$ hyperparameter, which controls the diversity of the generated text, was randomly selected to be anywhere from $p=0$ (argmax) to $p=1.0$ (full random sampling).

\section{Case Study}
To show the efficacy of \ROFT{} as an evaluation tool, we present a case study from our initial pilot of over 3000 annotations of generations from the news article domain.

\subsection{Data Collection}
While our eventual hope is for the \ROFT{} website to have enough organic traffic for useful data to be collected, for the purposes of this study, two hundred Amazon Mechanical Turk workers were paid to complete 10 annotations each on the website.
In total, we collected 3244 annotations (7.9\% of annotators continued past the minimum of 10 questions they were required to do to get paid).
10\% of examples the crowd workers saw were designated attention check questions in which the prompt explicitly stated they should select ``human-written" at every step.
About 25\% of crowd workers failed this check, and after filtering out these annotators, we were left with a total of 1848 high-quality annotations, which we will refer to as the filtered annotation set.

\subsection{Inter-Annotator Agreement}
There were 768 examples which had at least two crowd workers provide annotations for them (645 of which had at least three annotations provided).
This led to 6,115 instances of pairs of annotations on the same examples.
Of these, 18.3\% predicted the exact same sentence as the boundary, and 28.4\%, predicted boundaries at most one sentence apart from each other.
When considering only the filtered annotation set, there were 2,064 pairs of annotations.
Of these, 18.6\% predicted the exact same sentence as the boundary, and 28.3\% predicted boundaries at most one sentence apart from each other.

\subsection{Evaluation Measures}
We consider three methods for evaluating annotator ability.

\begin{figure}
    \centering
    \includegraphics[width=0.48\textwidth]{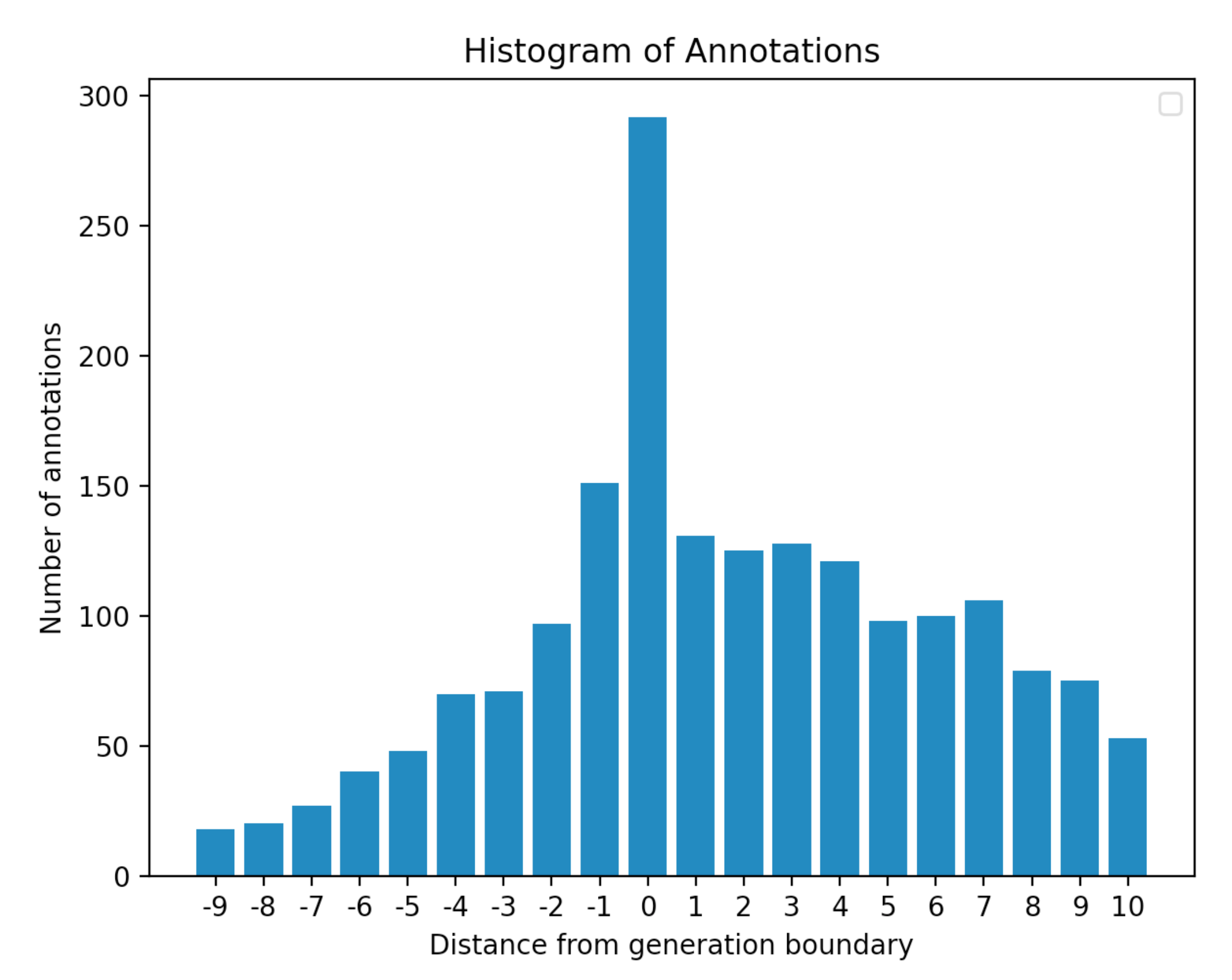}
    \caption{A histogram of the filtered annotation set grouped by the distance (in number of sentences) between the sentence selected by the annotator and the true boundary sentence.}
    \label{fig:tricklen}
\end{figure}

\subsubsection{Accuracy} Among annotators that passed our attention check, 15.8\% of the filtered annotations correctly identified the exact boundary between machine and generated text. Additionally, the average annotation from our filtered set was 1.989 sentences after the true boundary. This is consistent with our intuition, namely that current state-of-the-art NLG systems are capable of fooling humans but typically only for one or two sentences.

\subsubsection{Distance from Boundary} In Figure \ref{fig:tricklen}, we show a histogram of our filtered annotation set grouped by the distance each annotation was away from the true boundary.\footnote{As a note, values closer to zero in our histogram are more likely by construction as there are more opportunities for these distances to be selected. For example, a distance of -9 is only possible if the generation boundary is at the 10th sentence, while a distance of 0 is possible in every configuration. This does not affect our expectation that the distribution be symmetric if annotators are selecting at random.}
If annotators are selecting sentences at random, we would expect this distribution to be symmetric about 0. However, the observed distribution is significantly asymmetric, with the left tail (composed of annotators picking human-written sentences) dropping off precipitously while the right tail (composed of machine-generated sentences) decreases more linearly.
This asymmetry indicates that our annotators are successfully picking up on clues in the generated text, and thus the sentence-by-sentence structure of the \ROFT{} experiment is an effective way to evaluate text.
These preliminary results bode well for future large-scale use of the tool.

\subsubsection{Points Awarded} \label{points}
While accuracy may be a simple and intuitive metric for assessing performance, it is sub-optimal for our purposes as it does not give partial credit for guesses that are after the boundary, despite such guesses being successful identifications of generated text.
Average distance (in sentences) from boundary is not sufficient either, as it does not weight all guesses before the boundary equally negatively and thus over-penalizes too-early annotations on examples with late-occurring boundaries.

\begin{figure*}
\begin{subfigure}{.5\textwidth}
    \centering
    \includegraphics[width=0.8\textwidth]{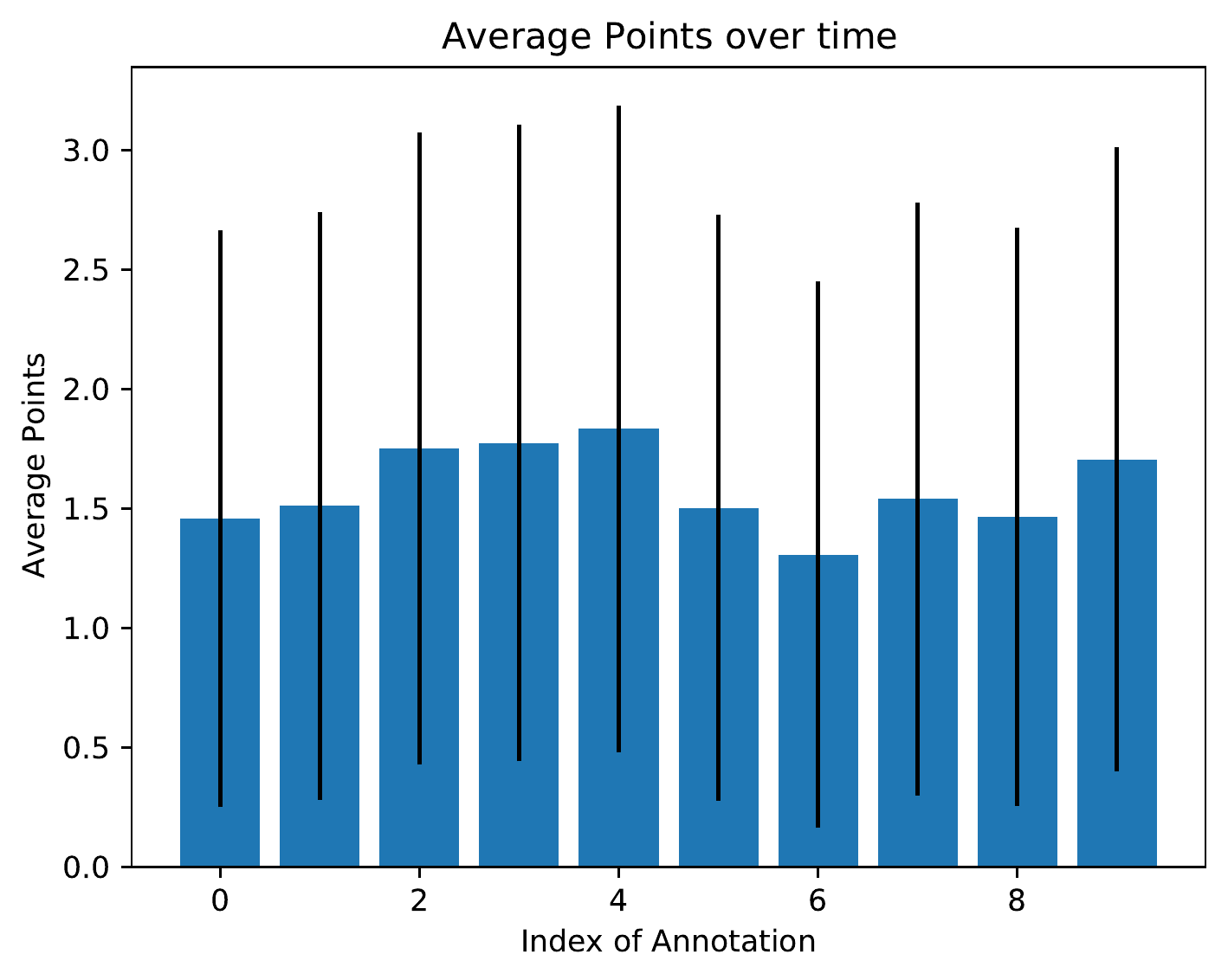}
    \caption{}
    \label{fig:overtime}
\end{subfigure}
\begin{subfigure}{.5\textwidth}
    \centering
    \includegraphics[width=0.8\textwidth]{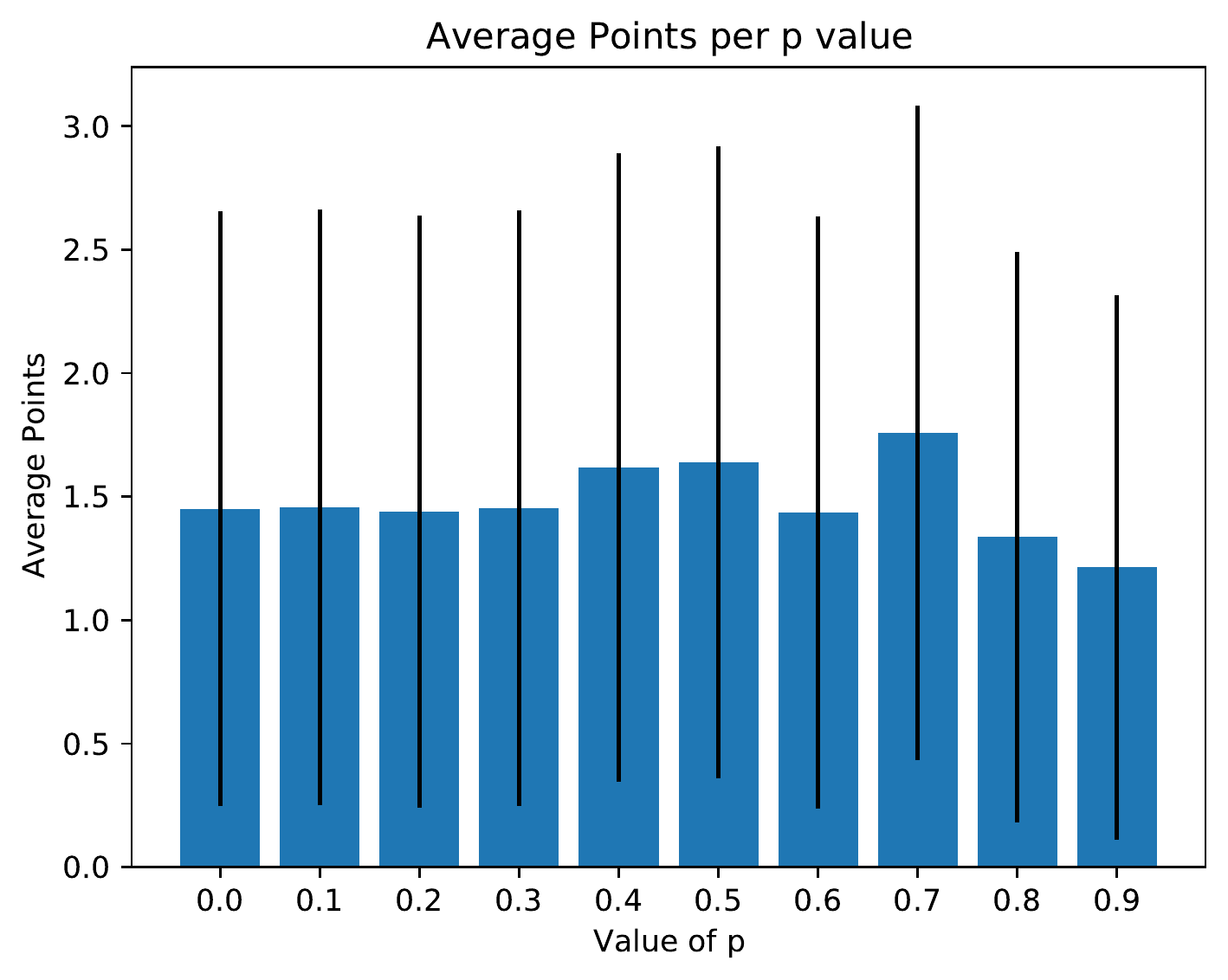}
    \caption{}
    \label{fig:boundary}
\end{subfigure}
\caption{In (a) we show the average number of points (Section 4.3) received per annotation in the filtered annotation set grouped by the temporal order in which they were shown to the annotators 0 (first) to 9 (last).
In (b) we show average number of points received per item in the filtered annotation for each values of $p$ used for decoding.
Error bars are standard deviation.
No statistically significant trends were observed in this preliminary study.}
\label{fig:points}
\end{figure*}

To combat these issues, we developed a point system to better capture annotator ability. After each annotation, a user is assigned points based on their performance: 5 points for guessing exactly on the boundary and a linearly decreasing number of points for each sentence beyond the boundary. No points are awarded for guesses that appear before the boundary. We use the average points per annotation as our metric for the experiments shown in Figure \ref{fig:points}.

\subsection{Skill Range of Annotators}
There was a significant range in detection ability across the crowd workers.
The top 5\% of the filtered worker pool earned an average of 3.34 points per annotations while the bottom 5\% earned an average of 0.35.
Since it is difficult to separate out the influence of inherent skill from that of misaligned incentives (AMT workers were paid for completion, not correctness), more research is necessary to understand differences in annotator ability.

\subsection{Impact of Decoding Strategy}
During our small-scale case study, we did not see a noticeable correlation between the values of the Nucleus Sampling \citep{holtzman2019curious} hyperparameter $p$ and the detection accuracy of humans as reported in Figure \ref{fig:boundary}. This is likely due to the low number of annotations per value of $p$ ($n$=180) and we hope to run a more comprehensive version of this experiment with more data in the future.

\subsection{Impact of Revealing the Boundary}
As part of the gamification aspect of the \ROFT{} platform, we reveal the true boundary to our annotators after every annotation they complete. This feature adds a level of interactivity to the process and is crucial for ensuring that the \ROFT{} experiment is enjoyable and appeals to the general public. To better understand how this decision affected annotator skill, we analyzed if our annotators got more accurate as they did more annotations. Figure \ref{fig:overtime} shows that over a session of 10 annotations, annotators exhibit little to no improvement at the annotation task over time. Future studies using the \ROFT{} platform will further investigate if human annotators can be trained to detect generated text over long periods of time and multiple gameplay sessions.

\subsection{Free-form Comments}
Our proposed annotation system allows annotators to provide a natural language explanation of why they made a particular decision (e.g. classifying a sentence as human-written or machine-generated). Due to minimal oversight, many annotators re-used or copy/pasted their comments across annotations. Filtering for duplicates, we collected over 1200 unique comments, out of around 3000 annotations. Manual inspection shows that many annotations relied on similar clues such as: problems with entailment, formatting (i.e. punctuation), and repetition. These responses can be used to inform future improvements to existing NLG systems and decoding strategies. Additionally, it is possible to use data mining techniques to extract an error taxonomy from the provided natural langauge description of errors.

\begin{table}[]
  \centering
  \small
 \begin{tabular}{| p{7cm} |}
     \hline
     \textbf{Sample Annotation} \\
     \hline\hline \\[-0.5em]
     Seems like a conversational statement that doesnt logically follow from a book title reference  \\ [1ex]
     \hline \\[-0.5em]
     not relevant to preceding sentences  \\ [1ex]
     \hline \\[-0.5em]
     I don't think that a human would write about tarot cards in an obituary and it says obituaries plural. \\ [1ex]
     \hline \\[-0.5em]
     The sentence is too short and simple, sweating computerized. \\ [1ex]
     \hline \\[-0.5em]
     First time I heard of dinosaur-eating mammals \\ [1ex]
     \hline \\[-0.5em]
     The sentence is left hanging. \\ [1ex]
     \hline \\[-0.5em]
     Repeated the second line again and To is written as TO \\ [1ex]
     \hline
    \end{tabular}
    \caption{Examples of explanations crowd workers gave for why they thought a sentence was machine-generated.}
    \label{tab:my_label}
\end{table}

\section{Related Work}
Nearly all papers in NLG do some form of human evaluation, usually using Amazon Mechanical Turk \citep{van2019best}.
Typically the interfaces for these evaluations are simple web forms.
\citet{van2019best} offers a survey of many of these methods.
Custom-designed websites for collecting or displaying human evaluations of generated text have become increasingly prominent in the open-ended dialog domain, with ChatEval \citep{sedoc-etal-2019-chateval} and ConvAI \citep{pavlopoulos2019convai} being two examples.

However, \ROFT{} was primarily influenced by other ``real or fake" websites that attempt to gamify the detection task, such as \url{http://www.whichfaceisreal.com/} for generated face images and \url{https://faketrump.ai/} for generated Tweets.
Our task is similar to the one used for human evaluation in \citet{ippolito2020automatic}, except in their task the text shown to raters was either entirely human-written or entirely machine-generated.

The boundary detection task we propose was inspired by the Dialog Breakdown Detection Challenge \citep{higashinaka2016dialogue}, in which the goal is to automatically detect the first system utterance in a conversation between a human and a chatbot system that causes a dialogue breakdown.

\section{Conclusion and Future Work}

In this work, we have introduced \ROFT{} and have shown how it can be used to collect annotations on how well human raters can tell when an article transitions from being human-written to being machine-generated.

Ultimately, we plan to use \ROFT{} to conduct a large-scale systematic study of the impact of decoding strategy, fine-tuning dataset, prompt genre, and other factors on the detectability of machine-generated text.
We also intend to collect and release a large dataset of natural language explanations for why humans think text is machine-generated.
We hope that these will provide insights into problems with the human-written text we use as prompts and into the types of errors that NLG systems make.

Such a study will require tens of thousands of human annotations.
We hope that by gamifying the annotation process and encouraging organic traffic to the website, we can ultimately bypass the need for crowd workers who, since they are paid by the annotation, are disincentivized from taking the time to provide high quality annotations.

We believe that \ROFT{} provides a powerful tool for understanding the strengths and limitations of a great variety of NLG systems, and we look forward to working with researchers interested in testing out their own model outputs within the \ROFT{} evaluation framework.

\section*{Acknowledgements}
This research is based upon work supported in part by the DARPA KAIROS Program (contract FA8750-19-2-1004), the DARPA LwLL Program (contract FA8750-19-2-0201), and the IARPA BETTER Program (contract 2019-19051600004). Approved for Public Release, Distribution Unlimited. The views and conclusions contained herein are those of the authors and should not be interpreted as necessarily representing the official policies, either expressed or implied, of DARPA, IARPA, or the U.S. Government.
The \ROFT{} website is also supported by a grant from the Google Cloud Platform research credits program.

We thank the members of our lab for their feedback on the design of the \ROFT{} user interface.

\bibliographystyle{acl_natbib}
\bibliography{roft}

\begin{thebibliography}{13}
\expandafter\ifx\csname natexlab\endcsname\relax\def\natexlab#1{#1}\fi

\bibitem[{Fan et~al.(2018)Fan, Lewis, and Dauphin}]{fan2018hierarchical}
Angela Fan, Mike Lewis, and Yann Dauphin. 2018.
\newblock Hierarchical neural story generation.
\newblock In \emph{Proceedings of the 56th Annual Meeting of the Association
  for Computational Linguistics (Volume 1: Long Papers)}, pages 889--898.

\bibitem[{Higashinaka et~al.(2016)Higashinaka, Funakoshi, Kobayashi, and
  Inaba}]{higashinaka2016dialogue}
Ryuichiro Higashinaka, Kotaro Funakoshi, Yuka Kobayashi, and Michimasa Inaba.
  2016.
\newblock The dialogue breakdown detection challenge: Task description,
  datasets, and evaluation metrics.
\newblock In \emph{Proceedings of the Tenth International Conference on
  Language Resources and Evaluation (LREC'16)}, pages 3146--3150.

\bibitem[{Holtzman et~al.(2019)Holtzman, Buys, Du, Forbes, and
  Choi}]{holtzman2019curious}
Ari Holtzman, Jan Buys, Li~Du, Maxwell Forbes, and Yejin Choi. 2019.
\newblock The curious case of neural text degeneration.
\newblock In \emph{International Conference on Learning Representations}.

\bibitem[{Ippolito et~al.(2020)Ippolito, Duckworth, Callison-Burch, and
  Eck}]{ippolito2020automatic}
Daphne Ippolito, Daniel Duckworth, Chris Callison-Burch, and Douglas Eck. 2020.
\newblock Automatic detection of generated text is easiest when humans are
  fooled.
\newblock In \emph{Proceedings of the 58th Annual Meeting of the Association
  for Computational Linguistics}, pages 1808--1822.

\bibitem[{Keskar et~al.(2019)Keskar, McCann, Varshney, Xiong, and
  Socher}]{keskar2019ctrl}
Nitish~Shirish Keskar, Bryan McCann, Lav~R Varshney, Caiming Xiong, and Richard
  Socher. 2019.
\newblock Ctrl: A conditional transformer language model for controllable
  generation.
\newblock \emph{SalesForce Einstein.ai blog}.

\bibitem[{van~der Lee et~al.(2019)van~der Lee, Gatt, van Miltenburg, Wubben,
  and Krahmer}]{van2019best}
Chris van~der Lee, Albert Gatt, Emiel van Miltenburg, Sander Wubben, and Emiel
  Krahmer. 2019.
\newblock Best practices for the human evaluation of automatically generated
  text.
\newblock In \emph{Proceedings of the 12th International Conference on Natural
  Language Generation}, pages 355--368.

\bibitem[{Liu et~al.(2018)Liu, Saleh, Pot, Goodrich, Sepassi, Kaiser, and
  Shazeer}]{liu2018generating}
Peter~J Liu, Mohammad Saleh, Etienne Pot, Ben Goodrich, Ryan Sepassi, Lukasz
  Kaiser, and Noam Shazeer. 2018.
\newblock Generating wikipedia by summarizing long sequences.
\newblock In \emph{International Conference on Learning Representations}.

\bibitem[{Pavlopoulos et~al.(2019)Pavlopoulos, Thain, Dixon, and
  Androutsopoulos}]{pavlopoulos2019convai}
John Pavlopoulos, Nithum Thain, Lucas Dixon, and Ion Androutsopoulos. 2019.
\newblock Convai at semeval-2019 task 6: Offensive language identification and
  categorization with perspective and bert.
\newblock In \emph{Proceedings of the 13th International Workshop on Semantic
  Evaluation}, pages 571--576.

\bibitem[{Radford et~al.(2019)Radford, Wu, Child, Luan, Amodei, and
  Sutskever}]{radford2019language}
Alec Radford, Jeffrey Wu, Rewon Child, David Luan, Dario Amodei, and Ilya
  Sutskever. 2019.
\newblock Language models are unsupervised multitask learners.
\newblock \emph{OpenAI Blog}, 1(8):9.

\bibitem[{Reiter and Belz(2009)}]{reiter2009investigation}
Ehud Reiter and Anja Belz. 2009.
\newblock An investigation into the validity of some metrics for automatically
  evaluating natural language generation systems.
\newblock \emph{Computational Linguistics}, 35(4):529--558.

\bibitem[{Sandhaus(2008)}]{sandhaus2008new}
Evan Sandhaus. 2008.
\newblock The new york times annotated corpus.
\newblock \emph{Linguistic Data Consortium, Philadelphia}, 6(12):e26752.

\bibitem[{Sedoc et~al.(2019)Sedoc, Ippolito, Kirubarajan, Thirani, Ungar, and
  Callison-Burch}]{sedoc-etal-2019-chateval}
Jo{\~a}o Sedoc, Daphne Ippolito, Arun Kirubarajan, Jai Thirani, Lyle Ungar, and
  Chris Callison-Burch. 2019.
\newblock \href {https://doi.org/10.18653/v1/N19-4011} {{C}hat{E}val: A tool
  for chatbot evaluation}.
\newblock In \emph{Proceedings of the 2019 Conference of the North {A}merican
  Chapter of the Association for Computational Linguistics (Demonstrations)},
  pages 60--65, Minneapolis, Minnesota. Association for Computational
  Linguistics.

\bibitem[{Zellers et~al.(2019)Zellers, Holtzman, Rashkin, Bisk, Farhadi,
  Roesner, and Choi}]{zellers2019defending}
Rowan Zellers, Ari Holtzman, Hannah Rashkin, Yonatan Bisk, Ali Farhadi,
  Franziska Roesner, and Yejin Choi. 2019.
\newblock Defending against neural fake news.
\newblock In \emph{Advances in Neural Information Processing Systems}, pages
  9054--9065.

\end{thebibliography}

\appendix
\end{document}